\documentclass[runningheads]{llncs}
\usepackage[T1]{fontenc}
\usepackage[hidelinks]{hyperref}
\usepackage{tikz}
\usepackage{wrapfig}
\usepackage{caption}
\usepackage{subcaption}
\usepackage{amsmath}
\usepackage{amssymb}
\usepackage{tabularx}
\usepackage{listings}
\usepackage{booktabs}
\usepackage{xcolor}
\usepackage{algorithm}
\usepackage{algorithmic}
\usepackage{mathrsfs}
\usetikzlibrary{automata, positioning, petri}

\usepackage{graphicx}
\begin{document}
\title{Discovering and Analyzing Stochastic Processes to Reduce Waste in Food Retail\thanks{The authors thank the support from Freiburg - Adelaide Partnership Fund.}}
%
\titlerunning{Discovering and Analyzing Processes to Reduce Waste in Food Retail}
%
\author{
Anna Kalenkova \inst{1}\orcidID{0000-0002-5088-7602}, 
Lu Xia \inst{1}\orcidID{0009-0000-0979-8053},\\
Dirk Neumann \inst{2}\orcidID{0000-0003-2178-3705}
}

\authorrunning{A. Kalenkova et al.}
%
\institute{School of Computer and Mathematical Sciences, \\ The University of Adelaide, Australia \\
\email{\{anna.kalenkova@,lu.xia@student.\}adelaide.edu.au}\\
 \and
  Information Systems Research,\\ Albert-Ludwigs-University of Freiburg, Germany\\ \email{dirk.neumann@is.uni-freiburg.de}
}

\maketitle              
\begin{abstract}
This paper proposes a novel method for analyzing food retail processes with a focus on reducing food waste. The approach integrates object-centric process mining (OCPM) with stochastic process discovery and analysis. First, a stochastic process in the form of a continuous-time Markov chain is discovered from grocery store sales data. This model is then extended with supply activities. Finally, a \emph{what-if} analysis is conducted to evaluate how the quantity of products in the store evolves over time. This enables the identification of an optimal balance between customer purchasing behavior and supply strategies, helping to prevent both food waste due to oversupply and product shortages.


\end{abstract}
\section{Introduction}
Globally, it is estimated that roughly one-third of all food produced for human consumption goes to waste~\cite{FAO2011}. This staggering figure represents more than a billion tons of food lost every year, exposing the inefficiencies of our current food system. Such waste occurs at various stages along the food life cycle: during harvest (due to poor storage or transport), in processing (where food may not meet quality standards), and at the consumer level (where oversupply may lead to unnecessary disposal). 
\emph{Food waste} is a significant sustainability challenge, contributing to the growing pressures on environmental and social systems.

Sustainability is achieved through the implementation of \emph{circular economy} principles, which aim to minimize waste, reduce resource consumption, and promote regenerative systems.
In~\cite{GEISSDOERFER2017757}, circular economy
is defined as ``\emph{a regenerative system in which resource input and waste, emission, and energy leakage are minimized by slowing, closing, and narrowing material and energy loops}''. While \emph{slowing} and \emph{closing} food retail loops refer to extending product use or enabling recycling and donations to charities, \emph{narrowing} primarily involves optimizing logistics to purchase only the amount of food likely to be sold~\cite{Bigdeloo2021}. In this paper, we study models that support the optimization of food supply (narrowing).

Since data describing retail processes often comes as a collection of sales events with timestamps and additional contextual information, process mining~\cite{vanDerAalst2016} emerges as a promising tool for analyzing these processes. It provides a comprehensive set of techniques~\cite{10292162} to support sustainability initiatives by enabling data-driven insights and process analysis. Although food waste reduction has been discussed within the realm of process mining~\cite{nikolov2015combining,phdowenkeates,Wuennenberg2023,Ting2014}, no specific process mining technique for food waste reduction has been proposed.

 In~\cite{Fritsch2025}, a 5 phase approach for process mining research strategy in sustainability area was proposed.
This paper focuses on the first three stages: modeling (discovery), analysis, and process improvement and optimization. We introduce methods for discovering stochastic processes that capture the dynamics of store capacity, specifically the amount of products on the shelves. We then provide analytical tools for performing what-if analysis on the discovered models, which can support process optimization and food waste reduction strategies.

The paper is organized as follows. \autoref{sec:lit} provides a literature review, including relevant research in process mining and, more broadly, in food waste analysis. The main event log and process model concepts are introduced in \autoref{sec:background}, followed by the presentation of process discovery and analysis techniques in~\autoref{sec:analysis}. The analysis of real-world data is presented in~\autoref{sec:case}. Finally, \autoref{sec:conclusion} concludes the paper.


\section{Literature Review}
\label{sec:lit}
This section provides an overview of relevant research on process mining, its applications in supply chain analysis, and existing methods for food waste analysis.

While no process mining techniques have been specifically developed for food waste analysis, several methods have been proposed for supply chain analysis more broadly. A comprehensive review of process mining techniques applicable to supply chains is presented in~\cite{Oldenburg2025}, and systematic literature reviews are available in~\cite{Jokonowo2018,Jacobi2020}. The work in~\cite{KNOLL2019130} bridges the gap between supply chain analysis and process mining by proposing concrete strategies for understanding and analyzing logistics processes. In~\cite{schatter2022supply}, process mining is applied to evaluate supply chain resilience. An overview of successful practical applications is provided in~\cite{Reil2021}, and a specific framework for analyzing food supply chains is introduced in~\cite{Keates2020}.

A key area of process mining suited for analyzing complex scenarios, such as supply chains involving multiple objects, is the so-called \emph{object-centric process mining (OCPM)}~\cite{vanderAalst2023}. Within the realm of OCPM, a trace in an event log can represent a sequence of events associated not with a single process case but with a specific object, such as a customer, item, or order~\cite{Adams2022}. The OCEL standard for object-centric event logs was developed and introduced in~\cite{Ghahfarokhi2021}. More recently, the application of OCPM and the OCEL standard to sustainability analyses was explored in~\cite{graves2025object}, where the authors propose a framework to support the reduction of negative impacts from companies' operations.

Another promising direction of OCPM that can advance sustainability analysis is \emph{stochastic OCPM}, which was introduced in~\cite{vanDetten2024}. This position paper emphasizes the importance of analyzing stochastic properties of object-centric models, which can enable the derivation of frequent process patterns. Although methods for discovering stochastic process models from event logs have been proposed in~\cite{RoggeSolti2014,Burke2021,Leemans2024,Alkhammash2024}, the research presented in this paper introduces a novel approach that integrates OCPM with stochastic process discovery and analysis.

A systematic literature review of approaches for (1) improving supply chain \emph{resilience}, i.e., the capacity of the supply chain to respond to interruptions, and (2) \emph{reducing food waste}, is presented in~\cite{Seyam2024}. For example, an optimization problem aiming to maximize profit and minimize total supply chain lead time was addressed in~\cite{Bottani2019}. A simulation method for analyzing supply chains under disruption was proposed in~\cite{Zhu2020}. Furthermore, a technique for analyzing supplier disruptions, which combines Markov chains to model supplier capacity and dynamic Bayesian networks to examine the propagation impact of a disruption, was presented in~\cite{Hosseini2019}.

An optimization problem that ensures food donations are proportional to each country’s demand while minimizing the amount of undistributed food was studied in~\cite{Orgut2016}. An optimal packaging problem that minimizes both transportation costs and food waste was proposed and applied to simulated data in~\cite{Heising2017}. An optimization model incorporating dynamic shelf life (early discounting of products) was studied and demonstrated its effectiveness in real-world settings~in~\cite{Buisman2019}.

Other food waste reduction approaches based on mathematical optimization techniques were proposed in~\cite{Ou2021,Kabadurmus2022}. Paper~\cite{Nikolicic2021} suggests simulation techniques to model and analyze waste in food supply chains. In~\cite{TorresSanchez2020}, a non-linear regression model relating temperature and maximum shelf life was introduced. A comprehensive analysis of various machine learning techniques for short-term demand forecasting in food catering services was proposed in~\cite{RODRIGUES2024140265}. An approach based on game theory that develops and investigates optimal food decision-making strategies between a retailer and a supplier was proposed in~\cite{Lin2023}. Research utilizing grey causal modeling to identify and analyze the main factors contributing to food waste was presented in~\cite{Ali2019,SINGH2023109172}.

In contrast to the above techniques, the approach presented in this paper leverages process mining and OCPM concepts to build \emph{stochastic process models} from event logs, enabling the modeling and analysis of food waste based on the temporal aspects of customer behavior.

\vspace{-5pt}
\section{Background}
\label{sec:background}
In this section, we define object-centric event logs and stochastic processes for application in food retail process discovery and analysis.

In retail, event logs capture various types of information, including transaction and product identifiers, timestamps, quantities of purchased products, payment methods, and customer types. This data can be  represented in a form of an object-centric event log that also contains timestamps of transactions. We will adapt the definitions provided in~\cite{Adams2022} and~\cite{10904251} to represent the retail processes event data.
\setcounter{footnote}{0}

\begin{definition}[Event log]
\label{def:event_log}
Let $\mathcal{E}$ be a universe of event identifiers, $\mathcal{O}$ a universe of objects, and $\mathcal{A}$, $\mathcal{V}$, and $\mathcal{T}$ the universes of attributes, values, and timestamps, respectively. An \emph{event log} $L=(E,O,f_o,f_a,f_t)$ is a tuple, where:
\begin{enumerate}
    \item $E\subseteq \mathcal{E}$ is a set of events;
    \item $O\subseteq \mathcal{O}$ is a set of objects;
    \item $f_o:E\rightarrow\mathcal{P}\footnote{$\mathcal{P}(O)$ denotes the powerset of $O$, i.e., the set of all subsets of $O$.}(O)$ is a function that maps events to subsets of objects;
    \item $f_a:E\times\mathcal{A}\rightharpoonup \mathcal{V}$ is a partial function that assigns values to some event attributes;
    \item $f_t:E\rightarrow \mathcal{T}$ defines the occurrence times of events.
\end{enumerate}
\end{definition}

An example of such an event log is presented in~\autoref{tab:event_log}. Each event from the set $E=\{e_1,e_2,e_3,e_4,e_5\}$ represents a transaction in which a client buys fruits in the grocery store. Objects are represented by fruits and clients: $O=\{orange, apple, mango, watermelon, client\ 1, client\ 2\}$. 

\begin{table}[t!] 
\centering
\footnotesize
\renewcommand\arraystretch{1.25}
\vspace{-10pt}
\caption{\label{tab:event-log}An event log of a grocery sales system.}
\label{tab:event_log}
\vspace{-10pt}
\begin{center}
\begin{tabular}{ l|c|c|c|c| } 
\cline{2-5}
                    & \textit{Objects} & \textit{Quantity}
                    & \textit{Total price}
                    & \textit{Timestamp} \\
\cline{2-5}
{\color{darkgray}$e_1$} & \{orange, client 1\} & 10 & 12.0 & 2025-07-19 08:23:42 \\
\cline{2-5}
{\color{darkgray}$e_2$} & \{apple, client 1\}  & 15 & 10.3 & 2025-07-19 08:24:03  \\
\cline{2-5}
{\color{darkgray}$e_3$} & \{orange, client 2\} & 5 & 6.0 & 2025-07-19 08:24:22  \\
\cline{2-5}
{\color{darkgray}$e_4$} &  \{mango, client 2\} & 2 & 8.0 & 2025-07-19 08:24:49  \\
\cline{2-5}
{\color{darkgray}$e_5$} & \{watermelon, client 2\} & 1 & 6.3 & 2025-07-19 08:25:05  \\
\cline{2-5}
\end{tabular}
\end{center}
\vspace{-20pt}
\end{table}
\normalsize

Events are mapped to sets of objects, as presented in~\autoref{tab:event_log}. For example, $f_o(e_1)=\{orange, client\ 1\}$. Attributes such as $\emph{Quantity}$ and $\emph{Total\ price}$ are assigned to all the events in the event log; for instance, $f_a(e_1,\emph{Quantity})=10$. Each event is also associated with a timestamp, e.g., $f_t(e_1)=\text{2025-07-19 08:23:42}$. Although in this event log   we consider not only products but also clients to make the definition applicable to different kinds of process analysis, including user behavior, the approach presented in the following sections focuses solely on products as objects.

The main type of process model used in our analysis is the so-called \emph{finite-state continuous-time Markov chain}, which we will refer to simply as a \emph{continuous-time Markov chain}.

\begin{definition}[Continuous-time Markov chain]
\label{def:markov-chain}
Let $S=\{0,1,\dots,k\}$ be a finite set of model states. A continuous-time Markov chain is defined as a pair: $\mathit{CTMC}=(\lambda,Q)$, where:
\begin{enumerate}
\item $\lambda=(p_0,p_1,\dots,p_k)$ is a probability vector on $S$ that defines the \emph{initial state probability}, such that $\sum\limits_{i\in S} p_i=1$.
\item $Q=(q_{i,j})_{i,j\in S}$ is a \emph{rate matrix} on $S$, such that: 1) $\forall i,j\in S, i\neq j, q_{i,j}\geq 0$; 2)  $\forall i\in S$ it holds that $\sum\limits_{j\in S} q_{i,j}=0$.

\end{enumerate}

\end{definition}

\begin{center}
\begin{figure}
\begin{tikzpicture}[>=latex, shorten >=1pt, auto, node distance=2.2cm,
                    semithick, every state/.style={circle, draw, minimum size=0.8cm, inner sep=1pt, font=\small}]
  \node[state] (0)                    {$k$};
  \node[state] (1) [right of=0]       {$k\!-\!1$};
  \node[state] (2) [right of=1]       {$k\!-\!2$};
  \node[state, draw=none] (dots) [right of=2] {$\cdots$};
  \node[state] (n) [right of=dots]    {$1$};
   \node[state] (dots2) [right of=n] {$0$};

  \path[->] (0) edge[bend left=20] node[above] {\small $q_{k,k\!-\!1}$} (1)
            (1) edge[bend left=20] node[above] {\small $q_{k\!-\!1,k\!-\!2}$} (2)
            (2) edge[bend left=20] node[above] {} (dots)
            (dots) edge[bend left=20] node[above] {} (n);

  \path[->]
            (2) edge[bend left=20] node[below] {\small $q_{k\!-\!2,k\!-\!1}$} (1);

 \path[<-] (0) edge[bend right=50] node[above] {\small $q_{k\!-\!2,k}$} (2)
 (n) edge[bend right=40] node[above] {\small $q_{k\!-\!2,1}$} (2)
             (1) edge[bend right=47] node[above] {\small $q_{1,k\!-\!1}$} (n);

  \path[<-] (2) edge[bend right=50] node[above] {\small $q_{k,k\!-\!2}$} (0)
  (dots2) edge[bend right=20] node[above] {$q_{1,0}$} (n);

\end{tikzpicture}
\caption{A continuous-time Markov chain with states $S=\{0,1,\dots,k\}$.}
\label{fig:ctmc_example}
\end{figure}
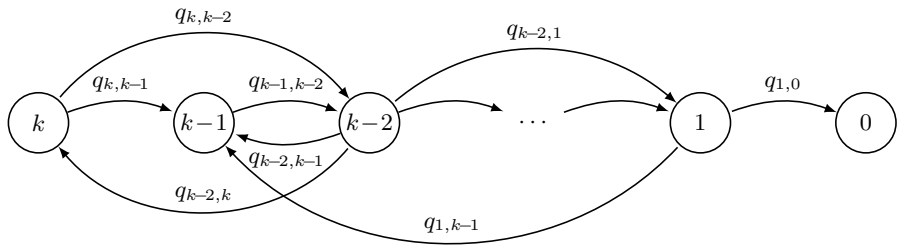
\end{center}

At each moment in time, the stochastic process defined by a continuous-time Markov chain is in one of the states from the set~$S$. Initial state of the process is selected randomly according to the vector $\lambda$.
When the process enters a state $i \in S$, the next state is determined based on the outgoing transition rates~$q_{i,j}$.
For each transition rate $q_{i,j} > 0$, the transition time is drawn from the exponential distribution with rate~$q_{i,j}$ with the probability density function $q_{i,j} e^{-q_{i,j} t}$.
The process then transitions to the state~$j$ with the minimal transition time, after waiting for that time delay.

A continuous-time Markov chain, presented in~\autoref{fig:ctmc_example}, consists of states and transitions between these states with positive rates. Notice that continuous-time Markov chains do not have self-loops (transitions connecting the state to itself), because the holding time in state $i$ is exponentially distributed with the rate~$\sum\limits_{j,j\neq i}q_{i,j}$. 

In this paper, we use the concept of irreducible continuous-time Markov chains, defined below.

\begin{definition}[Irreducible  continuous-time Markov chain]
Let $S$ be a set of states, and let $\mathit{CTMC} = (\lambda, Q)$ be a continuous-time Markov chain defined on this set.  
The chain $\mathit{CTMC}$ is called \emph{irreducible} if, for every pair of distinct states $i, j \in S$, there exists a sequence of transitions with positive rates that leads from $i$ to $j$.  

Formally, $\mathit{CTMC}$ is irreducible if:
$
\forall i, j \in S,\ i \neq j,\ \exists\ s_1, \dots, s_m \in S, \text{such that }\allowbreak s_1 = i,\ s_m = j,\ \text{and } q_{s_l, s_{l+1}} \text{ is positive for all } 1 \leq l < m.$
\end{definition}

\section{Discovery and Analysis of Stochastic Processes}
\label{sec:analysis}
This section presents an approach for the discovery and analysis of continuous-time Markov chains to model and analyze grocery store processes.
\subsection{Mining customer purchasing behavior from event data}
Consider an event log of a grocery store system (\autoref{tab:event_log}). First, this event log $L=(E,O,f_o,f_a,f_t)$ is filtered, and a sublog for each product is extracted. For example, for the product $\mathit{orange}$, the sublog $L'$ will contain only events that have $\mathit{orange}$ as an associated object. More formally, $L'=(E',O,f'_o,f'_a,f'_t)$, where $E'=\{e\in E|\mathit{orange}\in f_o(e)\}$, and $f'_o$,$f'_a$,$f'_t$ are the restrictions of functions  $f_o$,$f_a$,$f_t$ to $E'$, respectively. After the event log \(L'\) for a particular product has been derived, the corresponding continuous-time Markov chain will be discovered by applying Algorithm~1. Note that, in real-world store event data, each product typically corresponds to a single transaction (i.e., one event), and thus attributes such as quantity usually refer to only one product.

\begin{algorithm}
\caption{Discovery of a continuous-time Markov chain from event log \(L'=(E',O,f'_o,f'_a,f'_t)\)}
\begin{algorithmic}
\STATE \textbf{Input:} Event log \(L'\) for a specific product, maximum quantity of the product \(k\) (capacity), and initial quantity of the product \(i\).
\STATE \textbf{Output:} Continuous-time Markov chain \(\mathit{CTMC} = (\lambda, Q)\) on a set of states \(S\).
\begin{itemize}
    \item Define the set of states as \( S = \{0, \ldots, k\} \). 
    \item Set the initial state probability vector as \( \lambda = (0, \ldots, 0, 1, 0, \ldots, 0) \), such that \( p_j = 0 \) if \( j \neq i \), and \( p_i = 1 \).
    \item Collect all the purchased product quantities: $\mathcal{Q}=\{ f'_a(e',\emph{Quantity}) \mid e' \in E' \}$.
    \item For each $\scriptsize{\mathscr{Q}}\in\mathcal{Q}:$
    \subitem Filter the event log $L'$ to retrieve events with the corresponding quantity $\scriptsize{\mathscr{Q}}$, $E'_{\scriptsize{\mathscr{Q}}}=\{e'\in E'\mid f'_a(e',\emph{Quantity})= {\scriptsize{\mathscr{Q}}}\}$.
    \subitem If the size of the set $E'_{\scriptsize{\mathscr{Q}}}$ is 1 or all events have identical timestamps defined by the function $f'_a$, continue with the next quantity $\scriptsize{\mathscr{Q}}$ from $\mathcal{Q}$.
    \subitem Order the events from $E'_{\scriptsize\mathscr{Q}}$ into a sequence $l'_{\scriptsize\mathscr{Q}} = \langle e'_1, e'_2, \ldots, e'_m \rangle$ according to their timestamps, as defined by the function $f'_t$. Events with identical timestamps can be placed in any order.
    \subitem Calculate the time intervals for the sequence $l'_{\scriptstyle\mathscr{Q}}$ as a multiset\footnotemark of numbers: $\Delta'_{\scriptstyle\mathscr{Q}} = [f'_t(e'_{i+1}) - f'_t(e'_i) \mid 1 \leq i < m]$.
    \subitem Calculate the mean interval value $\mu'_{\scriptstyle\mathscr{Q}}$ for the multiset $\Delta'_{\scriptstyle\mathscr{Q}}$.
    \subitem For each state $i$, such that\footnotemark, ${\mathscr{Q}}\leq i\leq k$, set the  $q_{i,i\!-\!{\mathscr{Q}}}=\frac{1}{\mu'_{\scriptstyle\mathscr{Q}}}$.
    \item Set all undefined non-diagonal entries of the matrix $Q$ to 0 and define all diagonal entries as $q_{i,i}=-\sum\limits_{j,j\neq i}q_{i,j}$.
    \item Return \(\mathit{CTMC} = (\lambda, Q)\).
\end{itemize}
\end{algorithmic}
\end{algorithm}
\footnotetext[2]{Note that the same time interval may appear multiple times.}
\footnotetext[3]{We assume that $k$ is large enough and larger than any value of ${\mathscr{Q}}$ presented in the~data.}

This algorithm illustrates how to construct a model of clients' purchasing behavior for each product available in the store. A fragment of a purchasing behavior model, in which clients buy either one or two units of the product, is presented in~\autoref{fig:purchase}. Notice that in this model $q_{k,k\!-\!1}=q_{k\!-\!1,k\!-\!2}=q_{k\!-\!2,k\!-\!3}$, and 
 $q_{k,k\!-\!2}=q_{k\!-\!1,k\!-\!3}$.
The models discovered from event data using Algorithm~1 represent only the purchasing behavior. The following subsection discusses the enhancement of these models with supply transitions and their further analysis. 
 
\begin{center}
\begin{figure}
\vspace{-20pt}
\centering
\begin{tikzpicture}[>=latex, shorten >=1pt, auto, node distance=2.2cm,
                    semithick, every state/.style={circle, draw, minimum size=0.8cm, inner sep=1pt, font=\small}]
  \node[state] (0)                    {$k$};
  \node[state] (1) [right of=0]       {$k\!-\!1$};
  \node[state] (2) [right of=1]       {$k\!-\!2$};
  \node[state] (dots) [right of=2] {$k\!-\!3$};
  \node[state, draw=none] (n) [right of=dots]    {$\cdots$};
  \node[state, draw=none] (n2) [right of=n]    {$\cdots$};

  \path[->] (0) edge[bend left=20] node[above] {\small $q_{k,k\!-\!1}$} (1)
            (1) edge[bend left=20] node[above] {\small $q_{k\!-\!1,k\!-\!2}$} (2)
             (2) edge[bend left=50] node[above] {} (n)
            (2) edge[bend left=20] node[above] {$q_{k\!-\!2,k\!-\!3}$} (dots)
            (1) edge[bend left=50] node[above] {$q_{k\!-\!1,k\!-\!3}$} (dots)
            (dots) edge[bend left=20] node[above] {} (n)
            (dots) edge[bend left=50] node[above] {} (n2);
  \path[<-] (2) edge[bend right=50] node[above] {\small $q_{k,k\!-\!2}$}(0);

\end{tikzpicture}
\caption{A fragment of a continuous-time Markov chain modeling clients' purchasing behavior when clients buy either 1 or 2 units of the given product.}
\label{fig:purchase}
\end{figure}
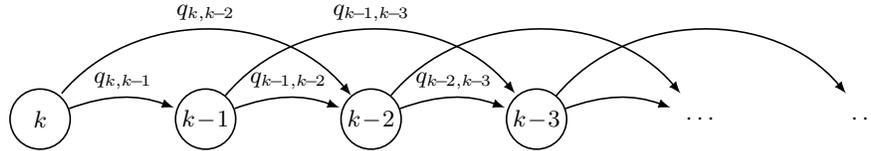
\end{center}

\subsection{Enhancement and analysis of the discovered continuous-time Markov chain}
Once the clients' behavior process is discovered, it can be enhanced with backward transitions that model the store’s supply strategies. For example, the store may choose to supply all products at the same rate, approximately once a week, with deliveries made in large batches to replenish the stock. This enhancement updates all models corresponding to different products by adding backward transitions with a specified supply rate \( q_s \), connecting states based on the quantity of supplied products. For example, if a product is supplied in batches of \( \mathscr{Q} \) units, each state \( 0 \leq i \leq k - \mathscr{Q} \), where \( k \) is the store’s capacity for the product, is connected to the state \( j = i + \mathscr{Q} \). The model assumes that no supply occurs if the resulting quantity would exceed the capacity \( k \).

The enhanced continuous-time Markov chain will contain both forward and backward transitions (as illustrated in~\autoref{fig:ctmc_example}) and should be checked for irreducibility. As will be demonstrated later, in real-life processes, almost all products can be sold in quantities of 1. Consequently, the resulting continuous-time Markov chain will contain forward transitions connecting neighboring states, such as \( i \) and \( i+1 \). We will show that, in this case, any supply strategy will result in an irreducible model.

\begin{theorem}[Irreducibility]
Let \( \mathit{CTMC} \) be a continuous-time Markov chain defined on a set of states \( S=\{0,1,\ldots,k\} \) and discovered by Algorithm~1 from store event data, where products are sold in quantities that include 1. Then, for any supply strategy that adds backward transitions between reachable states without exceeding the capacity, the resulting enhanced model is irreducible.
\end{theorem}

\begin{proof}
To prove irreducibility, we need to demonstrate that every state \( i \in S \) is reachable from any other state \( j \in S \). 
It suffices to show that the state corresponding to the maximum capacity \( k \) is reachable from \( j \), and that every state \( i \) is reachable from \( k \) via a sequence of forward transitions through the states \( k, k\!-\!1, \ldots, i \). 

Let \( \mathscr{Q} \) be the supply batch size. Starting in state \( j \), we can apply a sequence of backward transitions (representing supply events) to reach some state \( j' \) such that \( k\!-\!\mathscr{Q} < j' \leq k \). 

If \( j' = k \), we have reached the maximum capacity. Otherwise, we apply forward transitions from \( j' \) to \( k\!-\!\mathscr{Q} \), followed by a single backward transition to reach state \( k \). \qed

\end{proof}

Once an irreducible continuous-time Markov chain has been constructed, it can be used to analyze the model and determine the distribution over states, i.e., the probabilities of being in each state during a sufficiently long system run. These probabilities are called \emph{steady-state probabilities} and can be computed using the global balance equation: \( \pi Q = 0 \), where \( \pi = (\pi_1, \ldots, \pi_k) \) is the vector of steady-state probabilities, subject to the normalization condition \( \sum\limits_{1 \leq i \leq k} \pi_i = 1 \).

Knowing the probability of being in each state, that is, the probability of having a given number of product units in the store, allows for the assessment of food waste in relation to product oversupply. This analytical approach will be discussed and applied to real-world event data in the following section.

\section{Case Studies}
The proposed approach was implemented\footnote{The code is available at \url{https://github.com/akalenkova/foodwaste}.} and tested on real-world grocery store data\footnote{\url{https://www.kaggle.com/datasets/abhinayasaravanan/grocery-supply-chain-isuue}.}. The dataset contains grocery store transactions, including information on which product was purchased, in what quantity, and at what time. Overall, the dataset contains information on 300 unique product identifiers and 7,829 transactions. This dataset can be considered as an event log (\autoref{def:event_log}) to which Algorithm~1 can be applied. Because nearly all products had at least one transaction with a quantity of 1 (with only 8 exceptions), irreducible continuous-time Markov chains were constructed and analyzed.

\begin{figure}[h!]
  \centering
  \begin{subfigure}[b]{0.45\textwidth}
    \includegraphics[width=\linewidth]{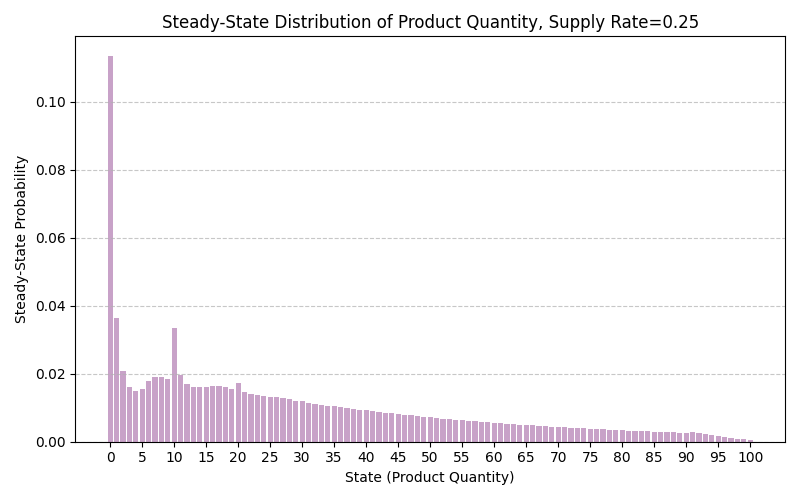}
    \caption{Steady-state distribution for a supply rate of 0.25 per hour.}
  \end{subfigure}
  \hfill
  \begin{subfigure}[b]{0.45\textwidth}
    \includegraphics[width=\linewidth]{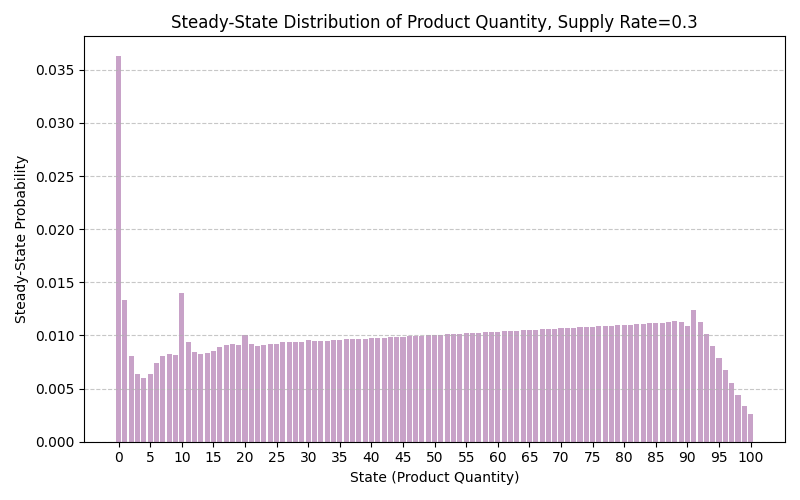}
    \caption{Steady-state distribution for a supply rate of 0.30 per hour.}
  \end{subfigure}

  \vspace{0.5cm}

  \begin{subfigure}[b]{0.45\textwidth}
    \includegraphics[width=\linewidth]{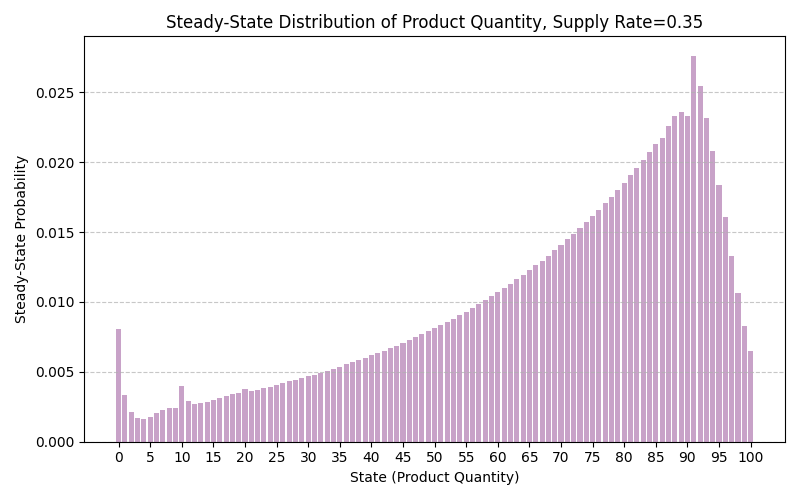}
    \caption{Steady-state distribution for a supply rate of 0.35 per hour.}
  \end{subfigure}
  \hfill
  \begin{subfigure}[b]{0.45\textwidth}
    \includegraphics[width=\linewidth]{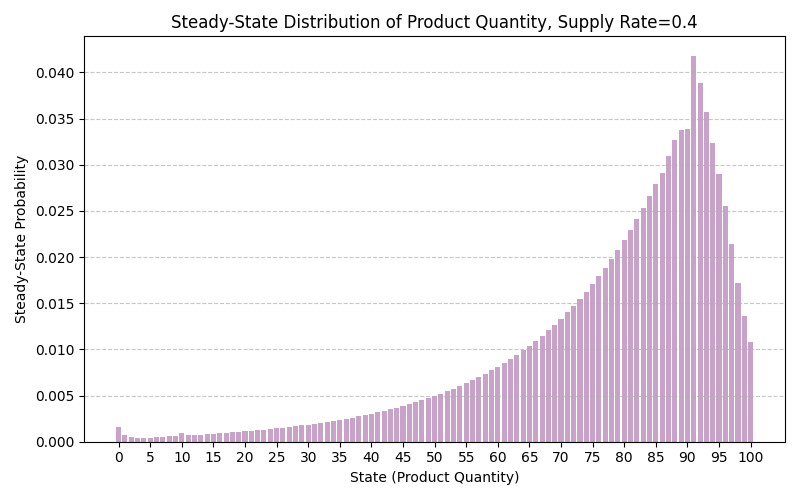}
    \caption{Steady-state distribution for a supply rate of 0.40 per hour.}
  \end{subfigure}

  \caption{Comparison of steady-state distributions across different supply strategies for a fruit product.}
  \label{fig:comparison}
\end{figure}

\autoref{fig:comparison} presents the steady-state distributions for continuous-time Markov chains constructed for a specific product from the fruit category. This fruit was purchased in quantities ranging from 1 to 4. The store capacity for this type of product was set to 100, and the batch size for the supply strategies was defined as 10. The supply rates considered included 0.25 (i.e., one supply every 4 hours), 0.30, 0.35, and 0.40. The probabilities of being in a state from 0 to 3 (indicating undersupply) are 0.1867, 0.0642, 0.0153, and 0.0031 for supply rates of 0.25, 0.30, 0.35, and 0.40, respectively. State 0 exhibits a relatively high steady-state probability. This is due to the fact that it only has backward transitions associated with supply, and no further purchases can occur from this state.

\label{sec:case}

Using the steady-state distributions, the expected product quantities corresponding to supply rates of 0.25, 0.30, 0.35, and 0.40 are calculated as 27.77, 49.35, 67.34, and 77.31, respectively. One may also observe that state 91 has a relatively high probability, which can be explained by the fact that it has no outgoing supply transitions due to the batch size of 10.

Assuming that any quantity exceeding a certain threshold (e.g., 70 units) results in food waste, the expected surplus can be calculated for each model. Using 70 as the threshold for the given product, the expected surplus, potentially indicating food waste, is 1.0239, 4.1024, 8.4592, and 12.1044 for supply rates of 0.25, 0.30, 0.35, and 0.40, respectively. The stochastic models highlight a trade-off: as the supply rate grows, undersupply diminishes, whereas oversupply~intensifies.

We consider this framework as a foundation for future developments in food waste analysis. First, we plan to develop methods for optimizing supply processes across all products in the store, recognizing that while supply batch sizes may vary, supply rates are often similar due to simultaneous delivery. Second, we aim to extend the continuous-time Markov chain model by explicitly incorporating food waste and linking it to product quantity. Finally, we intend to study transition times and generalize the model to accommodate non-exponential time distributions.

\section{Conclusion}
\label{sec:conclusion}
This paper combines object-centric process mining and stochastic process discovery to propose a novel approach for analyzing food retail processes. The proposed \emph{what-if} analysis enables the evaluation of different supply strategies by assessing the trade-off between food waste and product availability. By modeling customer purchase behavior with continuous-time Markov chains and enhancing these models with supply dynamics, we provide a quantitative basis for assessing and improving inventory management in grocery stores.

The approach was implemented and tested on real-world grocery store data, demonstrating its practical applicability. Through steady-state analysis, we were able to derive insights into stock levels, undersupply, and potential food waste under varying supply scenarios. Our results show that the method can effectively identify critical supply thresholds and offer guidance for decision-making in retail~logistics.


\bibliography{main.bib}

\begin{thebibliography}{10}
\providecommand{\url}[1]{\texttt{#1}}
\providecommand{\urlprefix}{URL }
\providecommand{\doi}[1]{https://doi.org/#1}

\bibitem{vanDerAalst2016}
van~der Aalst, W.M.P.: Process Mining: Data Science in Action. Springer, 2nd
  edn. (2016)

\bibitem{vanderAalst2023}
van~der Aalst, W.M.P.: Object-Centric Process Mining: An Introduction, pp.
  73--105. Springer International Publishing, Cham (2023)

\bibitem{Adams2022}
Adams, J.N., Schuster, D., Schmitz, S., Schuh, G., van~der Aalst, W.M.P.:
  Defining cases and variants for object-centric event data. In: Proceedings of
  the 4th International Conference on Process Mining (ICPM 2022). pp. 128--135.
  IEEE (2022)

\bibitem{Ali2019}
Ali, S.M., Moktadir, M.A., Kabir, G., Chakma, J., Rumi, M.J.U., Islam, M.T.:
  Framework for evaluating risks in food supply chain: Implications in food
  wastage reduction. Journal of Cleaner Production  \textbf{228},  786--800
  (2019)

\bibitem{Alkhammash2024}
Alkhammash, H., Polyvyanyy, A., Moffat, A.: Stochastic directly-follows process
  discovery using grammatical inference. In: Advanced Information Systems
  Engineering -- CAiSE 2024. Lecture Notes in Computer Science, vol. 14663, pp.
  87--103. Springer (2024)

\bibitem{Bigdeloo2021}
Bigdeloo, M., Teymourian, T., Kowsari, E., Ramakrishna, S., Ehsani, A.:
  Sustainability and circular economy of food wastes: Waste reduction
  strategies, higher recycling methods, and improved valorization. Materials
  Circular Economy  \textbf{3}(3) (2021)

\bibitem{Bottani2019}
Bottani, E., Murino, T., Schiavo, M., Akkerman, R.: Resilient food supply chain
  design: Modelling framework and metaheuristic solution approach. Computers \&
  Industrial Engineering  \textbf{135},  177--198 (2019)

\bibitem{Buisman2019}
Buisman, M., Haijema, R., Bloemhof-Ruwaard, J.: Discounting and dynamic shelf
  life to reduce fresh food waste at retailers. International Journal of
  Production Economics  \textbf{209},  274--284 (2019)

\bibitem{Burke2021}
Burke, A.T., Leemans, S.J.J., Wynn, M.T.: Stochastic process discovery by
  weight estimation. In: Process Mining Workshops: ICPM 2020 International
  Workshops, Padua, Italy, October 5–8, 2020, Revised Selected Papers.
  Lecture Notes in Business Information Processing, vol.~406, pp. 260--272.
  Springer (2021)

\bibitem{vanDetten2024}
van Detten, J.N.: Stochastic object-centric process mining: Analysing object
  interaction patterns. In: Proceedings of the 6th International Conference on
  Process Mining (ICPM 2024). vol.~3783, pp. 199--211. CEUR Workshop
  Proceedings (2024)

\bibitem{FAO2011}
{Food and Agriculture Organization of the United Nations}: Global food losses
  and food waste: Extent, causes and prevention. Tech. rep., FAO, Rome, Italy
  (2011), \url{https://www.fao.org/3/mb060e/mb060e.pdf}, study conducted for
  the International Congress SAVE FOOD! at Interpack2011, Düsseldorf, Germany

\bibitem{Fritsch2025}
Fritsch, A., Ullrich, M., Graves, N., Klessascheck, F., Nurkasanah, I.: Process
  science for sustainability: Research gaps and research strategy. In: EMISA
  2025. pp. 119--130. Lecture Notes in Informatics (LNI), Gesellschaft für
  Informatik (2025)

\bibitem{GEISSDOERFER2017757}
Geissdoerfer, M., Savaget, P., Bocken, N.M., Hultink, E.J.: The circular
  economy – a new sustainability paradigm? Journal of Cleaner Production
  \textbf{143},  757--768 (2017)

\bibitem{Ghahfarokhi2021}
Ghahfarokhi, A.F., Park, G., Berti, A., van~der Aalst, W.M.P.: {OCEL}: A
  standard for object-centric event logs. In: New Trends in Databases and
  Information Systems (ADBIS 2021). Communications in Computer and Information
  Science, vol.~1450, pp. 169--175. Springer (2021)

\bibitem{graves2025object}
Graves, N., Fritsch, A., Hensen, R., Koren, I., van~der Aalst, W.M.P.:
  Object-centric process mining for semi-automated and multi-perspective
  sustainability analyses. In: Proceedings of the 2025 International Conference
  on ICT for Sustainability (ICT4S). IEEE (2025)

\bibitem{10292162}
Graves, N., Koren, I., van~der Aalst, W.M.: Rethink your processes! a review of
  process mining for sustainability. In: 2023 International Conference on ICT
  for Sustainability (ICT4S). pp. 164--175 (2023)

\bibitem{Heising2017}
Heising, J.K., Claassen, G.D.H., Dekker, M.: Options for reducing food waste by
  quality-controlled logistics using intelligent packaging along the supply
  chain. Food Additives \& Contaminants: Part A  \textbf{34}(10),  1672--1680
  (2017)

\bibitem{Hosseini2019}
Hosseini, S., Ivanov, D., Dolgui, A.: Ripple effect modelling of supplier
  disruption: Integrated {M}arkov chain and dynamic {B}ayesian network
  approach. International Journal of Production Research  \textbf{58}(11),
  3284--3303 (2019)

\bibitem{Jacobi2020}
Jacobi, C., Meier, M., Herborn, L., Furmans, K.: Maturity model for applying
  process mining in supply chains: Literature overview and practical
  implications. Logistics Journal: Proceedings  \textbf{2020}(12),  1--12
  (2020)

\bibitem{Jokonowo2018}
Jokonowo, B., Claes, J., Sarno, R., Rochimah, S.: Process mining in supply
  chains: A systematic literature review. International Journal of Electrical
  and Computer Engineering  \textbf{8}(6),  4626--4636 (2018)

\bibitem{Kabadurmus2022}
Kabadurmus, O., Kazançoğlu, Y., Yüksel, D., Özbiltekin Pala, M.: A circular
  food supply chain network model to reduce food waste. Annals of Operations
  Research  (2022)

\bibitem{10904251}
Kalenkova, A., Mitchell, L., Roughan, M.: Performance analysis: Discovering
  semi-markov models from event logs. IEEE Access  \textbf{13},  38035--38053
  (2025)

\bibitem{phdowenkeates}
Keates, O.: Advancing Process Analytics for Agri-food Supply Chains. Phd
  thesis, Queensland University of Technology (2023)

\bibitem{Keates2020}
Keates, O., Wynn, M.T., Bandara, W.: A multi perspective framework for enhanced
  supply chain analytics. In: Business Process Management: 18th International
  Conference, BPM 2020. Lecture Notes in Computer Science, vol. 12168, pp.
  489--504. Springer (2020)

\bibitem{KNOLL2019130}
Knoll, D., Reinhart, G., Prüglmeier, M.: Enabling value stream mapping for
  internal logistics using multidimensional process mining. Expert Systems with
  Applications  \textbf{124},  130--142 (2019)

\bibitem{Leemans2024}
Leemans, S.J.J., Li, T., Montali, M., Polyvyanyy, A.: Stochastic process
  discovery: Can it be done optimally? In: Advanced Information Systems
  Engineering -- CAiSE 2024. Lecture Notes in Computer Science, vol. 14663, pp.
  36--52. Springer (2024)

\bibitem{Lin2023}
Lin, S.W., Januardi: Two-stage pricing of perishable food supply chain with
  quality-keeping and waste reduction efforts. Managerial and Decision
  Economics  \textbf{44}(3),  1749--1766 (2023)

\bibitem{Nikolicic2021}
Nikoličić, S., Kilibarda, M., Maslarić, M., Mircetic, D., Bojic, S.:
  Reducing food waste in the retail supply chains by improving efficiency of
  logistics operations. Sustainability  \textbf{13}(12), ~6511 (2021)

\bibitem{nikolov2015combining}
Nikolov, B.: Combining Data Mining and Process Mining for Analyzing Food Safety
  Processes. Master's thesis, Eindhoven University of Technology (2015)

\bibitem{Oldenburg2025}
Oldenburg, F., Hoberg, K., Leopold, H.: Process mining in supply chain
  management: State-of-the-art, use cases and research outlook. International
  Journal of Production Research  \textbf{63}(8),  2889--2904 (2025)

\bibitem{Orgut2016}
Orgut, I.S., Ivy, J.S., Uzsoy, R., Wilson, J.R.: Modeling for the equitable and
  effective distribution of donated food under capacity constraints. IISE
  Transactions  \textbf{48}(3),  252--266 (2016)

\bibitem{Ou2021}
Ou, T.Y., Lin, G.Y., Liu, C.Y., Tsai, W.L.: Constructing a sustainable and
  dynamic promotion model for fresh foods based on a digital transformation
  framework. Sustainability  \textbf{13}(19),  10687 (2021)

\bibitem{Reil2021}
Reil, T., Groher, E., Siegfried, P.: Process mining in supply chain management.
  Supply Chain Management Journal  \textbf{12}(2) (2021)

\bibitem{RODRIGUES2024140265}
Rodrigues, M., Miguéis, V., Freitas, S., Machado, T.: Machine learning models
  for short-term demand forecasting in food catering services: A solution to
  reduce food waste. Journal of Cleaner Production  \textbf{435},  140265
  (2024)

\bibitem{RoggeSolti2014}
Rogge-Solti, A., van~der Aalst, W.M.P., Weske, M.: Discovering stochastic
  {P}etri nets with arbitrary delay distributions from event logs. In: Business
  Process Management Workshops: BPM 2013 International Workshops, Beijing,
  China, August 26, 2013, Revised Papers. Lecture Notes in Business Information
  Processing, vol.~171, pp. 15--27. Springer (2014)

\bibitem{schatter2022supply}
Schätter, F., Haas, F., Morelli, F.: Supply chain resilience management using
  process mining. In: Proceedings of the 36th ECMS International Conference on
  Modelling and Simulation. pp. 121--127. European Council for Modelling and
  Simulation (2022)

\bibitem{Seyam2024}
Seyam, A., Barachi, M.E., Zhang, C., Du, B., Shen, J., Mathew, S.S.: Enhancing
  resilience and reducing waste in food supply chains: a systematic review and
  future directions leveraging emerging technologies. International Journal of
  Logistics Research and Applications pp. 1--35 (2024)

\bibitem{SINGH2023109172}
Singh, G., Rajesh, R., Daultani, Y., Misra, S.C.: Resilience and sustainability
  enhancements in food supply chains using digital twin technology: A grey
  causal modelling (gcm) approach. Computers \& Industrial Engineering
  \textbf{179},  109172 (2023)

\bibitem{Ting2014}
Ting, S.L., Tse, Y.K., Ho, G.T.S., Chung, S.H., Pang, G.: Mining logistics data
  to assure the quality in a sustainable food supply chain: A case in the red
  wine industry. International Journal of Production Economics  \textbf{152},
  200--209 (2014)

\bibitem{TorresSanchez2020}
Torres-Sánchez, R., Martínez-Zafra, M.T., Castillejo, N., Guillamón-Frutos,
  A., Artés-Hernández, F.: Real-time monitoring system for shelf life
  estimation of fruit and vegetables. Sensors  \textbf{20}(7), ~1860 (2020)

\bibitem{Wuennenberg2023}
Wuennenberg, M., Wegerich, B., Fottner, J.: Towards data management and data
  science for internal logistics systems using process mining and
  discrete-event simulation. Procedia CIRP  \textbf{120},  852--857 (2023)

\bibitem{Zhu2020}
Zhu, Q., Krikke, H.: Managing a sustainable and resilient perishable food
  supply chain (pfsc) after an outbreak. Sustainability  \textbf{12}(12), ~5004
  (2020)

\end{thebibliography}

\end{document}